\documentclass[10pt]{article}

\usepackage{amsmath,amssymb}
\usepackage{lingmacros}
\usepackage{url}

\usepackage[table]{xcolor}
\usepackage{array}
\usepackage{placeins}
\usepackage{multirow}

\usepackage{graphicx}

\usepackage[superscript]{cite}
\usepackage{multibib}

\newcites{Main}{References}
\newcites{Method}{Method References}

\begin{document}
\vspace*{0.2in}

\begin{flushleft}
{\Large
\textbf{\newline{Large Language Models can impersonate politicians and other public figures}}
}\newline
\\

Steffen Herbold, Alexander Trautsch, Zlata Kikteva, Annette Hautli-Janisz

\bigskip
Faculty of Computer Science and Mathematics, University of Passau\\
Dr.-Hans-Kapfinger-Str. 30\\
94032 Passau, Germany
\bigskip

Corresponding author: steffen.herbold@uni-passau.de

\end{flushleft}

\section*{Abstract}
Modern AI technology like Large language models (LLMs) has the potential to pollute the public information sphere with made-up content, which poses a significant threat to the cohesion of societies at large. 
A wide range of research has shown that LLMs are capable of generating text of impressive quality, including persuasive political speech, text with a pre-defined style, and role-specific content. 
But there is a crucial gap in the literature: We lack large-scale and systematic studies of how capable LLMs are in impersonating political and societal representatives and how the general public judges these impersonations in terms of authenticity, relevance and coherence. 
We present the results of a study based on a cross-section of British society that shows that LLMs are able to generate responses to debate questions that were part of a broadcast political debate programme in the UK. The impersonated responses are judged to be more authentic and relevant than the original responses given by people who were impersonated. 
This shows two things: (1) LLMs can be made to contribute meaningfully to the public political debate and (2) there is a dire need to inform the general public of the potential harm this can have on society. 
\section*{Introduction}

Modern Artificial Intelligence (AI) like ChatGPT~\citeMain{openai2023gpt4}, Claude~\citeMain{anthropic2024claude}, and Gemini~\citeMain{team2023gemini} is able to support humanity by generating high-quality textual content including source code~\citeMain{ziegler2024measuring}, persuasive student essays~\citeMain{Herbold2023}, and legal analysis~\citeMain{doi:10.1098/rsta.2023.0254}. More questionable in terms of advancement is the fact that LLMs seem (at least partly) to be able to mimic the linguistic pattern of authors \citeMain{bhandarkar-etal-2024-emulating}, to generate content that reflects general political identity~\citeMain{simmons2022moral, hackenburg2023comparing} and to assume the role of domain experts and answering as if they are in the role of an (unspecified) person from that domain~\citeMain{salewski2024context}. Downright questionable in terms of whether this technology advances humankind and promotes societal cohesion is the fact that LLMs are able to influence the political opinion of humans~\citeMain{Bai2023}, purport political bias \citeMain{santurkar2023opinionslanguagemodelsreflect} and successfully generate targeted persuasive communication  \citeMain{matz2024potential, simchon2024persuasive}. 
Yet another step towards poisoning the processes that shape public opinion is developing technology that can be used to impersonate public figures and elected political representatives. This is the starting point of the present paper. 

In our study, we condition an LLM in such a way that it impersonates a specific figure in the political and societal sphere in the UK. We then request from it a response to a question that the actual person has been confronted with in a debate on national television. Based on a representative cross-section of British society (n=948), we study how UK citizens rate the actual response of the person in comparison to the impersonated response along three axes: authenticity (the likelihood that the impersonated response comes from the actual person), coherence (the logical flow of the response), and relevance (the extent to which the response is relevant to the question). We also elicit the citizens' openness to using AI technology in public debates, both before and after engaging with the data of the study. This allows us to address two research questions:


\begin{description}
    \item[RQ1]: To what extent do UK citizens rate the authenticity, relevance, and coherence of impersonated debate responses differently from actual debate responses by the person?
    \item[RQ2]: What is the general public's view on using AI in public debates and is this view affected by exposure to technology?
\end{description}

The questions and responses underlying our study originate from thirty episodes of BBC's Question Time from 2020 to 2022 \citeMain{hautli-janisz-etal-2022-qt30}, one of the most viewed political debate programmes in the UK. The public figures in the dataset belong to one of six categories:  politicians (50\%),  business people (16.67\%),  journalists (14.17\%), medical experts (6.67\%),  writers (5.83\%), and other well-known members of society (activists, actors, political expert and sports personality -- 6.67\%). 
The survey participants are asked to (i) attribute both actual and impersonated response to public persons; (ii) evaluate how coherent and relevant both actual and impersonated responses are; and (iii) express their opinion regarding the use of AI in public debates. The last task is split into subtasks: First, the participants give their opinions of AI unaware of the source of the material they just rated (actual vs.~impersonated), then they are shown the source and they express their opinion on the technology again.

\section*{Results}

\subsection*{Impersonation is credible}

\begin{figure}
    \centering
    \includegraphics[width=\textwidth]{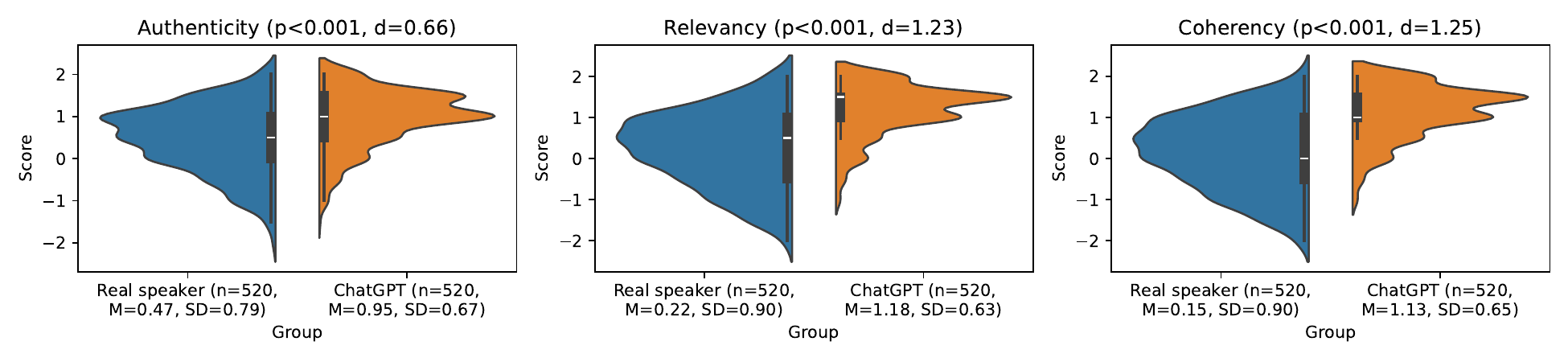}
    \caption{Judgments when a debate question, the name of the speaker, and either the ChatGPT-generated or the response by the actual speaker were shown. Violins show a kernel density estimation of the probability distribution, the miniature box-plots depict the median, upper and lower quartile, and the whiskers the largest/smallest value observed within 1.5 times the interquartile range of the upper/lower quartile. The statistical markers reported are the the p-value of two-sided Wilcoxon signed rank tests, the effect size with Cohen's $d$, the sample sizes $n$, mean values $M$ and standard deviations $SD$.}
    \label{fig:track-1}
\end{figure}

\begin{figure}
    \centering
    \includegraphics[width=\textwidth]{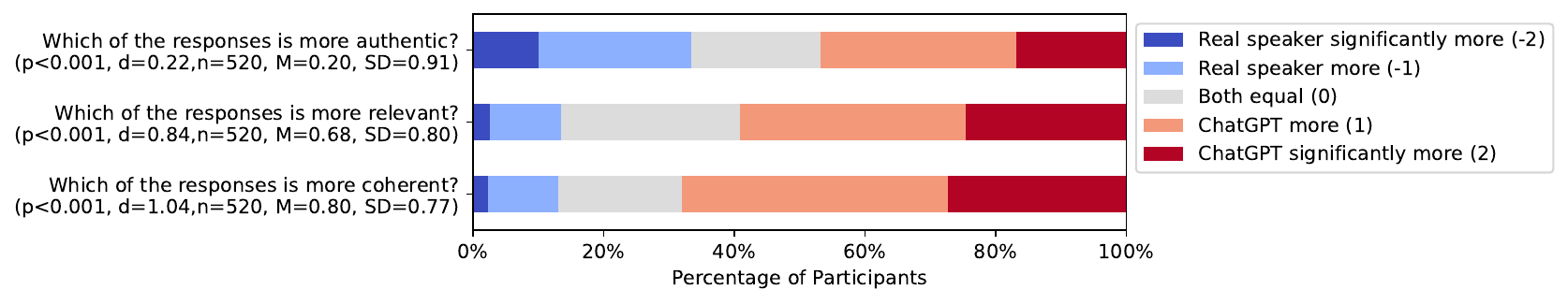}
    \caption{Judgments when a debate question, the name of the speaker, and both the actual and ChatGPT-generated responses were shown side-by-side. The stacked bar chart reports the percentages of the ratings that we observed. The statistical markers reported are the the p-value of a two-sided one-sample Wilcoxon signed rank tests for a difference from zero, the effect size with Cohen's $d$, the sample sizes $n$, mean values $M$ and standard deviations $SD$.}
    \label{fig:track-2}
\end{figure}

\begin{figure}
    \centering
    \includegraphics[width=\textwidth]{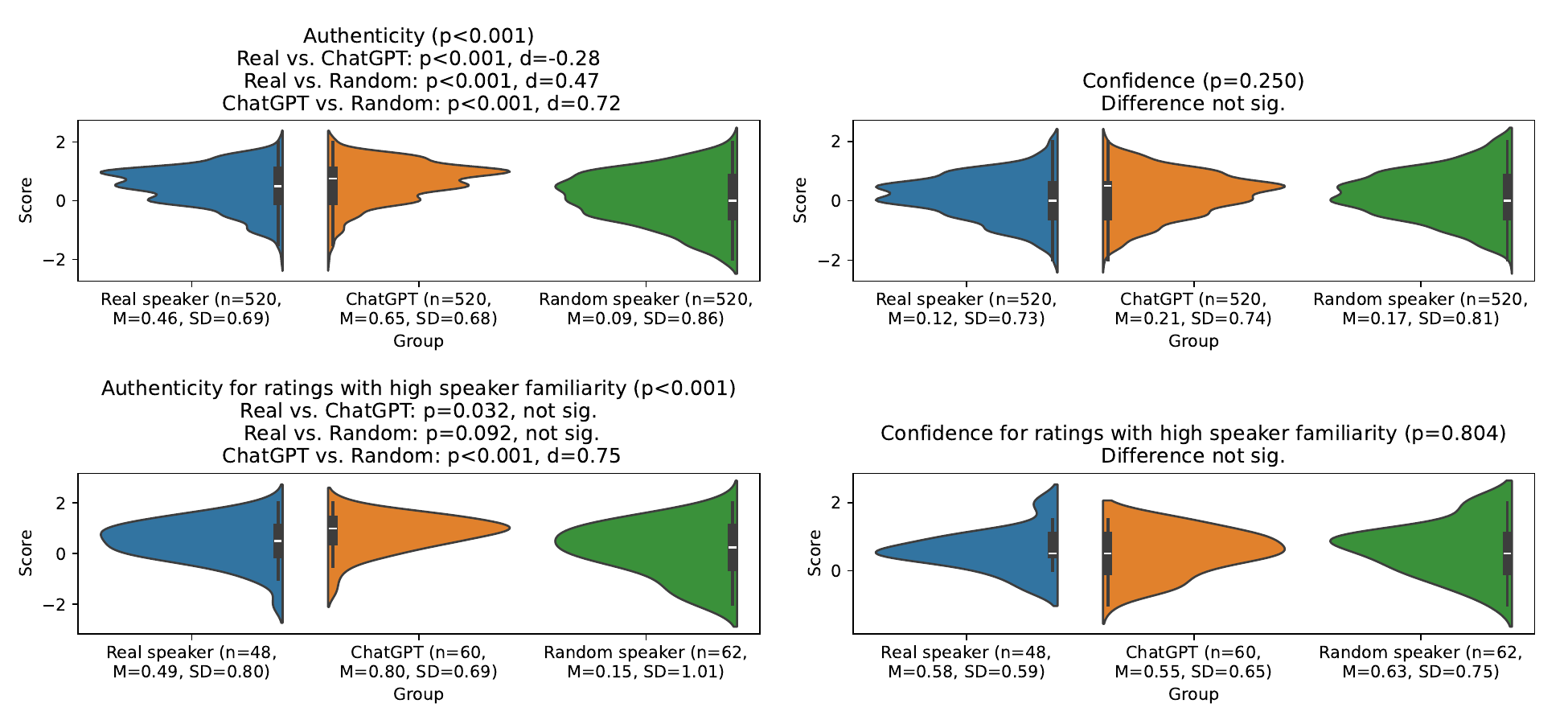}
    \caption{Judgments when a debate question with either the response and biography from the actual speaker, the ChatGPT-generated response and the biography of the actual speaker, or the response from the actual speaker but the name and biography of a random public person were shown. Violins show a kernel density estimation of the probability distribution, the miniature box-plots depict the median, upper and lower quartile, and the whiskers the largest/smallest value observed within 1.5 times the interquartile range of the upper/lower quartile. The statistical markers reported are the the p-value of the omnibus test for differences and pair-wise Bonfferoni-Dunn correct two-sided post-hoc tests, the effect size with Cohen's $d$, the sample sizes $n$, mean values $M$ and standard deviations $SD$.}
    \label{fig:track-3}
\end{figure}

Our results show clearly that LLM-generated, impersonated content is judged as more authentic, coherent, and relevant than the actual debate responses. When we only show one question and its response (either actual or impersonated) (see Figure~\ref{fig:track-1}), we observe a significant difference across all three dimensions with a medium effect size ($d=0.66$) for authenticity and a large effect size for coherence ($d=1.25$) and relevance ($d=1.23$) of the responses. When the participants directly compare an impersonated response with an actual response (see Figure~\ref{fig:track-2}) along these dimensions, the results are supported: The effect sizes for  relevance ($d=0.84$) and coherence ($d=1.04$) remain large, but the difference in authenticity decreases to a small effect ($d=0.22$), so the gap between impersonated responses over actual responses is smaller in this setting (but still significant). The effect size is comparable ($d=0.28$) if the participants see the biographies of the speakers during rating (see Figure~\ref{fig:track-3}, top-left) -- here the authenticity of impersonated response is still higher than that of the actual response.

\begin{figure}[h]
    \centering
    \includegraphics[width=\textwidth]{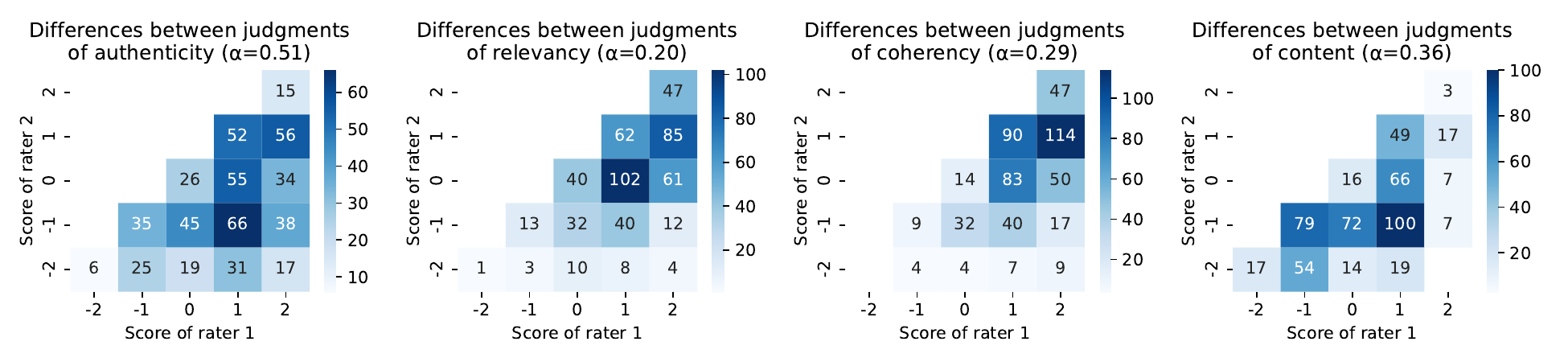}
    \caption{Inter-rater reliability when a debate question, the name of the speaker, and both the actual and Chat-GPTgenerated response were shown side-by-side. The heatmap visualizes the counts of the different rating combinations. The reported statistical marker is Cronbach's $\alpha$.}
    \label{fig:track-3-reliability}
\end{figure}

An important control for our study is whether transcribed debate content is just generally assumed to be authentic, instead of only when the response matches the common knowledge that the public has about the speaker. The data in Figure~\ref{fig:track-3-reliability} shows that when deliberately mis-assigning an actual response to a random speaker, the authenticity is significantly lower in comparison to when the response is assigned to the actual speaker ($d=0.46$) and the impersonated response ($d=0.71$). When we take into account the confidence of the participants in their rating, we do not find any significant differences in certainty of attributing an actual response or an impersonated response to an actual speaker, or a attributing the actual response to random speaker. If we only consider a subgroup of data where the participants are highly familiar with the speakers, this again neither affects the authenticity nor the confidence: While the significance tests for differences between actual and impersonated responses, as well as actual speakers versus random speakers are not significant anymore, the distributions look almost exactly the same as for the full dataset. This indicates that there is no shift in distributions, but rather that the effects are too small to be detected with the smaller subgroups.

\subsection*{Content is different}

\begin{figure}
    \centering
    \includegraphics[width=\textwidth]{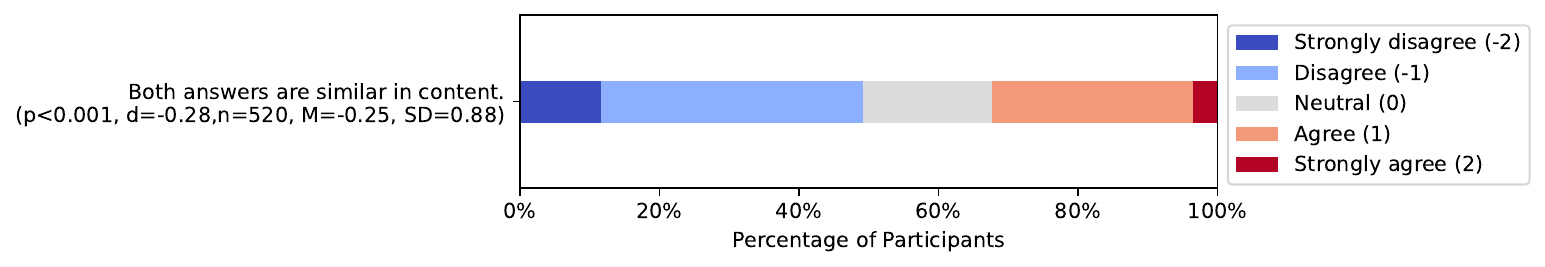}
    \caption{Judgments whether the content of the actual response and the ChatGPT-generated response are the same. The actla and impersonated response where shown side-by-side. The stacked bar chart reports the percentages of the ratings that we observed. The statistical markers reported are the the p-value of a two-sided one-sample Wilcoxon signed rank tests for a difference from zero, the effect size with Cohen's $d$, the sample sizes $n$, mean values $M$ and standard deviations $SD$.}
    \label{fig:track-2-content}
\end{figure}

A factor that should influence the participants' judgments of the authenticity of an impersonated response is its similarity in content to the actual response. If the content is different, i.e. the impersonated content is not in line with actual statements of the person, and authenticity is still rated high, we are faced with the situation where AI technology can used for targeted misinformation about the speaker's point of view. Our results show that a significant majority of actual responses is judged to be different in content to the impersonated counterpart, though the spread in the distribution is fairly large (see Figure~\ref{fig:track-2-content}). About half of the responses are considered to be dissimilar, in comparison to only about one third of the responses that are considered similar. Moreover, we observe no notable pattern or correlation between the similarity of the content and the authenticity of the responses ($\rho=-0.16$).

\subsection*{Linguistic structure is different}

To provide a perspective different from the human evaluation, we augment our results with an analysis of the linguistic surface of the responses (see Figure~\ref{fig:lingustic_surface}). This computational linguistic analysis, which is largely comparable to a previous approach of measuring the linguistic properties of LLM-generated text, sheds light on a number of aspects: First, the complexity of the sentences in terms of the number of conjuncts, clausal modifiers, clausal complements, clausal subjects and parataxes as well as the use of modals such as `should' and `must' are not significantly different between the actual responses and the impersonated ones. Secondly, the actual responses contain more discourse markers (e.g., `because', `therefore') than the impersonated responses, even though with a small effect size ($d=-0.36$). The reason for the statistically significant difference is that there is a long tail of actual responses that contain many discourse markers, even though the peak of the distributions is the same for actual and impersonated responses. Moreover, the actual speakers use substantially more epistemic markers like `I think' in their responses -- these expressions are only rarely found in the impersonated statements, leading to a large effect size ($d=-1.05$). Thirdly, the impersonated responses contain more nominalizations ($d=1.39$) and have a higher lexical diversity ($d=1.67$), both with a large effect size. The overlap between the words from the question and the response is higher for impersonated responses than actual responses, with a large effect size ($d=1.81$). In fact, the distribution shows that it is not uncommon that all words from the question appear in the impersonated responses, while this is only rarely so for the actual responses. 

\subsection*{Human judgement is reliable}

We use five-point Likert scales for the human judgments. In all variables, we observe an overall modest agreement when measured with Cronbach's $\alpha$, with values of at most $\alpha=0.55$. We analysed the data to understand which combination of different judgments we observed. 
Most notably, the data of the authenticity for all variants of the question shows that while there are differences in the judgments, there are relatively few polar differences, i.e., one participant rating an item as authentic and the other as not authentic. For relevance, coherence and content the differences are rather in how positive a judgment is with small differences of a single point (e.g., `neutral' instead of `agreement'), again showing that while the absolute ratings have some variance, the tendency regarding the judgment is typically the same for both participants. Overall, the tendency towards positive or negative judgements about a variable is fairly consistent, especially given our large sample size which is a representative cross-cut of the British society.

\subsection*{Public opinion}

\begin{figure}
    \centering
    \includegraphics[width=\textwidth]{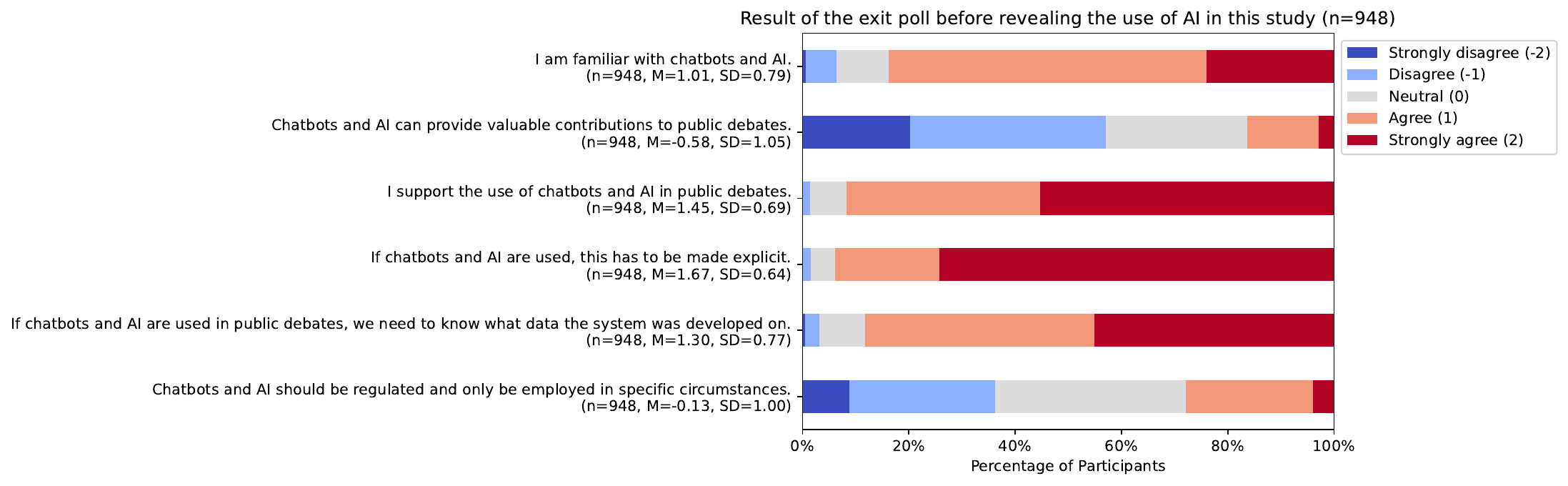}
    \includegraphics[width=0.8\textwidth]{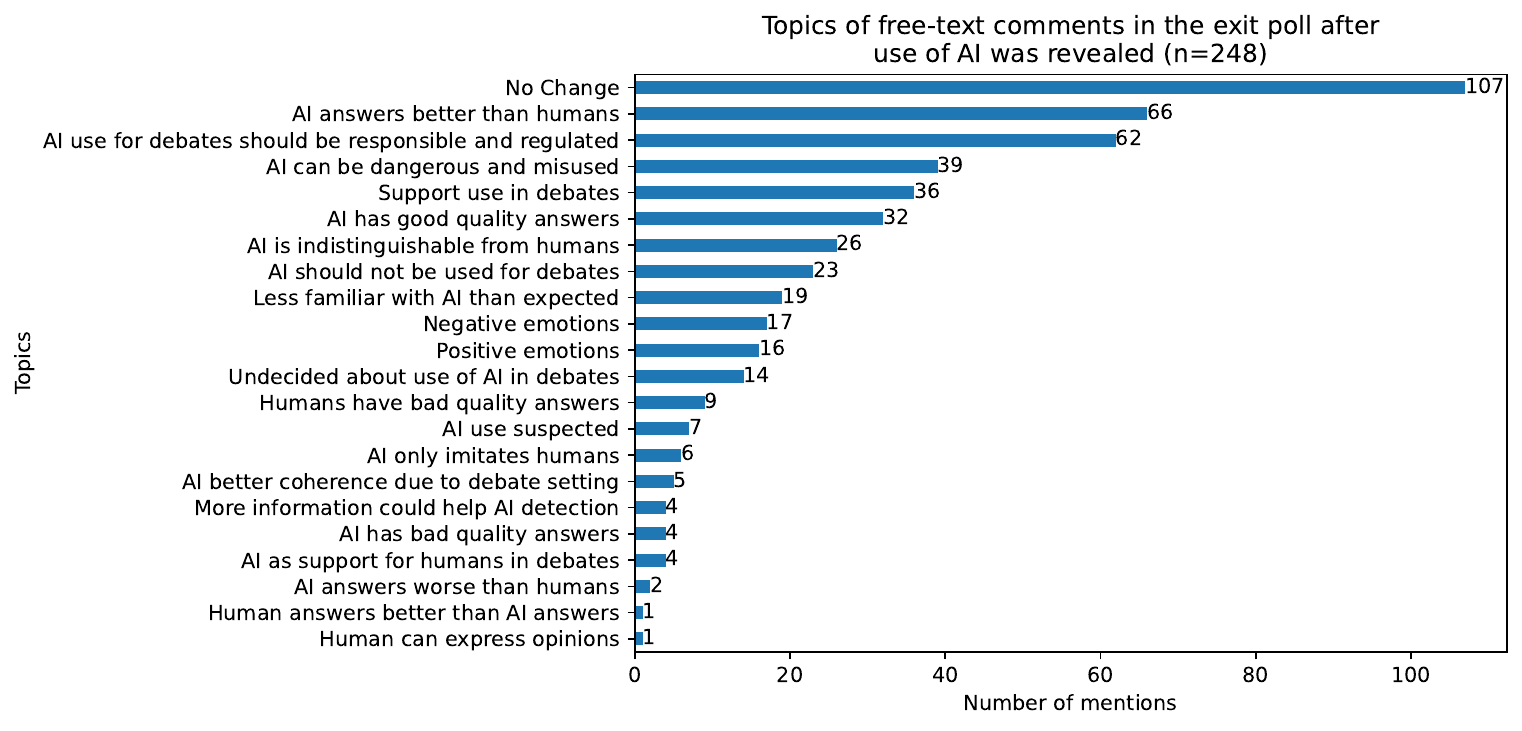}
    \caption{Results of the exit poll on the opinion of the participants. The stacked bar chart reports the percentages of the ratings that we observed. The statistical markers reported are the sample sizes $n$, mean values $M$ and standard deviations $SD$. The bar chat depicts the counts for each topic that was addressed in the free-text answers.}
    \label{fig:exit-poll-before}
\end{figure}

Public opinion on the use of AI technology for public debates was assessed in two steps: First, the participants gave their judgements without knowing the source of the data they had just rated in either task (i) or (ii), in the second step the source of the data was revealed and they were asked the same questions on AI technology again. Regarding the first step (see Figure~\ref{fig:exit-poll-before}) 
The results of our exit poll prior to revealing the use of AI paints a clear picture of the public opinion on the use of AI in public debates (see Figure~\ref{fig:exit-poll-before}). The participants mostly state that they are familiar with AI. Interestingly, while they mostly believe that AI cannot provide valuable contributions to public debates, they simultaneously state that they support the use of AI use, nevertheless, if it is made explicit and it is known how the system was developed. However, regarding a general regulation of AI, the participants provide a rather mixed picture, where there are roughly equal-sized groups favoring regulation, opposing regulation, and being undecided. Over 90\% of the participants did not change these opinions once we revealed the use of AI and asked if this affects their point of view. For those who changed their opinion, we found a clear trend: the participants realize they are less familiar with AI than they thought, but also have a better opinion on the use of AI in debates, while at the same time seeing a bigger need for regulation. 

The optional free-text answers ($n=248$) further corroborate these results. Many participants explicitly note that they did not change their results ($n=107$). However, the other free-text answers indicate that the changes in opinion are caused by the confrontation with the capabilities of AI through the survey. Participants often mention that the impersonated responses are better than the human responses ($n=66$) or that the quality of the impersonated responses is higher ($n=32$). A few participants noted that the high coherence in the impersonated responses made them sceptic towards AI use in the survey ($n=7$) and also that this advantage over the humans can be explained by the setting, where the humans do not have time to carefully prepare their responses ($n=5$). One participant even notes that this advantage of AI means that AI could be used to train humans for debates. Nevertheless, many also note that they are not able to distinguish between AI and humans at all ($n=26$). There are also a few comments noting negative aspects of the AI quality ($n=4$) or that AI was worse than the humans ($n=1$), but these are  rather outliers. 

Another aspect that is stressed in the comments is the requirement to regulate the use of AI ($n=62$), especially with respect to transparency: particularly the potential of deceptive use and the associated risks worry many participants ($n=39$), some even note feelings of fear, shock, and worry ($n=17$). However, some participants also express positive emotions like surprise and amazement given the strong capabilities of the AI ($n=16$). When it comes to the use in debates, some participants argue that the good performance shows a potential for use in debates ($n=36$), while others rather question the general concept AI debaters ($n=23$), for instance people question how AI can represent party opinions at all or what the actual worth of debate is without humans.

\section*{Discussion}

Our results demonstrate that \textbf{modern AI based on LLMs is able to provide high-quality impersonated debate content that is deemed authentic when attributed to actual people}. We also find indications that people rate the impersonated content to be slightly more authentic than the actual human debate responses. There does not seem to be a problem with an uncanny valley~\citeMain{mori1970bukimi} which makes humans feel uncomfortable with the impersonated answers. In addition, the impersonated responses are deemed more coherent and relevant than actual responses. While the lower coherence can be attributed to the humans being under scrutiny in a nationally broadcast TV politcal debate programme, the higher relevance of the LLM-generated responses indicates that the LLM stays better on-topic than the human speakers. 

Interestingly, we found that the authenticity is not negatively affected by the notable differences in the linguistic surface of the responses. The LLMs clearly had their own unique style marked marked by a diverse vocabulary and an avoidance of epistemic markers, but this was not picked up on by our participants. 

Even though most of \textbf{our participants stated that they are familiar with AI, they did not expect that AI could have generated these answers and underestimated the capabilities of modern generative AI}. When the participants were confronted with the strong capabilities of the AI, this elicited different responses: evidence-driven discussions of the merits of AI, including how to use it; negative emotional responses due to potential for misuse; and positive emotional responses due to the technological capabilities. This knowledge increases the appreciation for the capabilities of AI, but also the desire for regulation. 

When asked on the merits of AI, there is a clear belief that AI can be a valuable tool. There is no clear picture from the participants when asking for strong regulation and restrictions of use. However, when it comes to transparency, the public perspective is clear: \textbf{over 85\% of participants think that AI use has to be made explicit and that information on how the AI was developed needs to be shared}. 

The risks that are implied by our results are severe. We already know that LLMs are capable of generating persuasive misinformation\citeMain{10.1145/3544548.3581318} and that the automated and human detection of such misinformation is unreliable.\citeMain{doi:10.1137/1.9781611978032.50} Our work adds another layer on top of this: We demonstrate that LLMs can generate authentic information by impersonating specific people, meaning that LLM-powered misinformation campaigns can go beyond targeting general topics and target individual people by impersonating statements they contribute to the public discourse. 
Since the dissemination of excerpts from political statements via social networks is a common form of political communication~\citeMain{OSATUYI20132622}, it is easy to spread such generated statements. We have not yet tested how this works when we not only generate responses, but responses that push a specific political agenda. Content moderation to remove false generated statement is crucial~\citeMain{Clark2023}. 
Our own preliminary work suggests that a current model\citeMain{guo2023closechatgpthumanexperts} can be used for such content moderation (accuracy of 89\% on the task of classifying responses into impersonated or real). More sophisticated approaches may be able to fool such detectors~\citeMain{sun2024exploringdeceptivepowerllmgenerated}, but one can hope that this may be at the cost of authenticity.

Nevertheless, the implications of our results for the communication of political content are devastating: \textbf{threat actors can easily use LLMs to pollute public information spheres with fake but authentic political statements}, e.g., to sow confusion about what the actual remarks were and to invent talking points. If this is further combined with deep fakes that are already known to be able to generate reliable authentic voices and videos of public people~\citeMain{farid2022creating}, the potential for harm is enormous.

\FloatBarrier

\begin{figure}
    \centering
    \includegraphics[width=\textwidth]{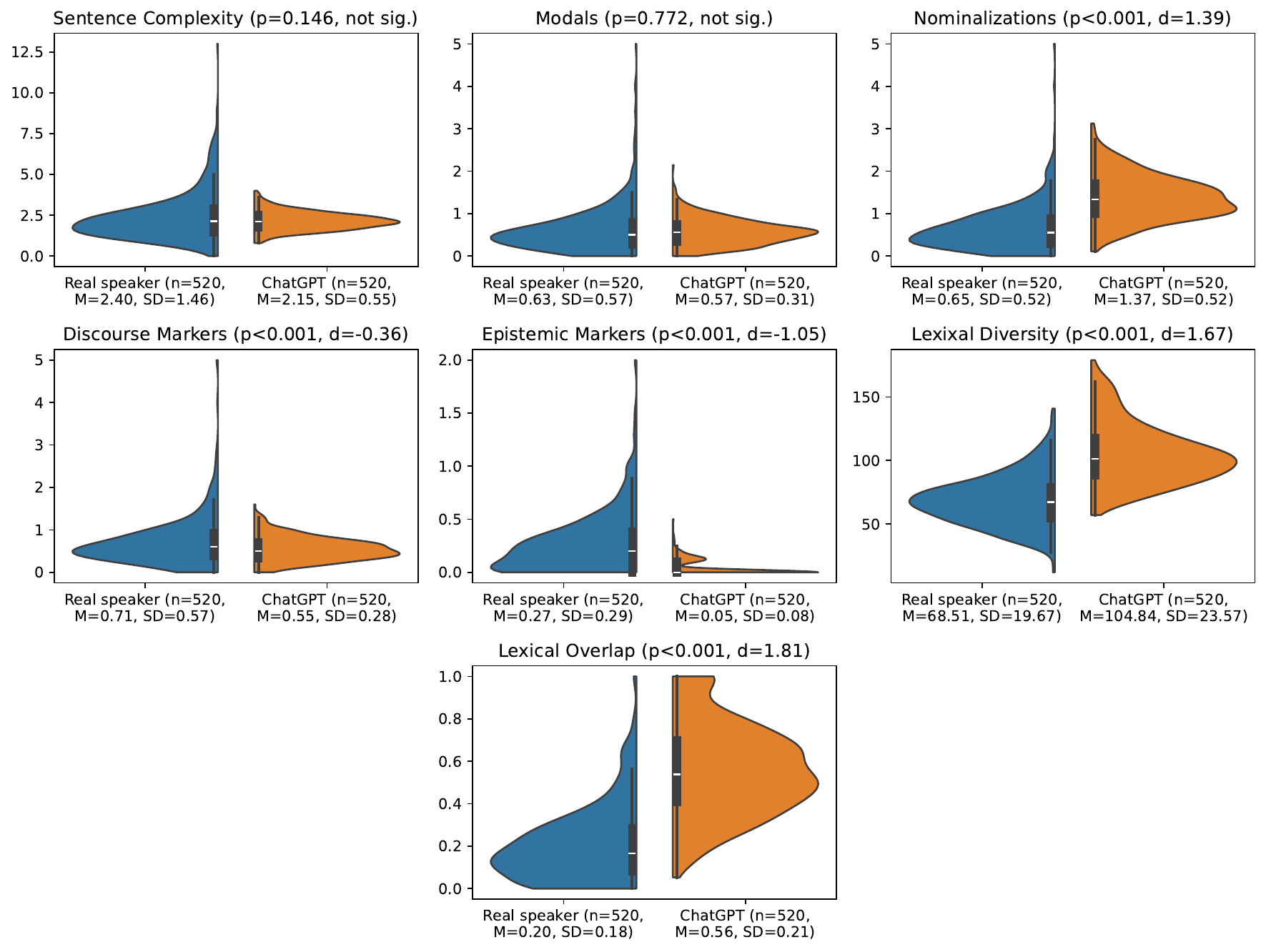}
    \caption{Linguistic surface of actual debate responses versus impersonated debate responses. Violins show a kernel density estimation of the probability distribution, the miniature box-plots depict the median, upper and lower quartile, and the whiskers the largest/smallest value observed within 1.5 times the interquartile range of the upper/lower quartile. The statistical markers reported are the the p-value of two-sided Wilcoxon signed rank tests, the effect size with Cohen's $d$, the sample sizes $n$, mean values $M$ and standard deviations $SD$.}
    \label{fig:lingustic_surface}
\end{figure}

\FloatBarrier

\bibliographystyleMain{unsrt}
\bibliographyMain{custom.bib}

\section*{Methods}

We hypothesize that modern LLMs are capable of generating political speech that is considered authentic for specific public persons, based on prior work that shows that political content can be generated~\citeMain{Bai2023}, linguistic styles imitated~\citeMain{bhandarkar-etal-2024-emulating}, and roles assumed~\citeMain{salewski2024context}. We measure this phenomenon based on the extraction of question/response pairs from a public debate corpus, the generation of new debate responses for these questions with a LLM, a survey on the human judgment of debate responses to measure the differences between the actual and impersonated responses, and an assessment of the linguistic surface of the impersonated responses. 

\subsection*{Real debate data}

The actual questions and responses are take from QT30 \citeMain{hautli-janisz-etal-2022-qt30}, currently the largest dataset of broadcast political debate. The corpus comprises the transcriptions of 30 episodes of the British talk show `Question Time' (QT) between June 2020 and November 2021. QT features a moderated panel format, driven by questions from the audience on the current topics of the week. The panelists are directed by the moderator to respond to the questions independent of a prior conversation on the topic and the initial statements by other panellists. We manually extract these questions and responses from the corpus and have a total of 119 unique questions with 555 responses from 119 different speakers. We discard the responses of seven speakers who did not have a Wikipedia page, which is requirement to generate debate responses (see below) and at the same serves as filter regarding whether the speakers are actually personalities in the public sphere. We also discard one response where the corpus data did not contain information about the speaker. This yields a set of 527 valid question/response pairs from 112 different speakers. We randomly drop seven responses to achieve a final count of 520 question/response pairs, because we require a sample size that is dividable by eight to get paired from participants that each judge eight question/response pairs.

\subsection*{Impersonated debate content}

We use a complex emulation protocol to impersonate people similar to Bhandarkar et al.~\citeMain{bhandarkar-etal-2024-emulating} with the following prompts:
\begin{itemize}
    \item System prompt: \texttt{You are an expert at mimicking different persons in debates. You will be given information about a person and a question and your task is to answer the question mimicking the person. You only answer as the person you are asked to mimic. Do not say the name of the person you are mimicking. Don't introduce yourself. Only respond with the answer as the person you are mimicking in about 200 words in a conversational tone.}
    \item User prompt: \texttt{Please only answer this question: [QUESTION] as this person: [SPEAKER\_WIKIPEDIA]. Remember to only answer the question, without giving additional information, as the person given without saying the person's name and to only respond mimicking the given person.}
\end{itemize}

The system prompt defines the behaviour we expect from the model, i.e., mimicking persons to impersonate them and to briefly answer questions in a conversational tone an introduction, as is common during debates. The user prompt starts with the task, then provides the question and a short biography of the speaker we obtain from the first paragraph of their Wikipedia article, as this paragraph provides a summary of information on their origin, party affiliation, political offices and so on. 
The user prompt then repeats the task to prompt the model to give the response in the expected format, followed by a manual sanity check to ensure that the impersonated responses are appropriate, i.e., do not contain the name of the speaker or information that the response was generated by a LLM, or a reason why no response was possible, e.g., due to lack of access to real-time data or for ethical reasons. This check did not flag problematic content, meaning that the LLM is able to generate appropriate responses for all debate questions. These we use in the subsequent study. 

As LLM, we used ChatGPT 4 Turbo~\citeMain{openai2023gpt4}. While more recent models, e.g., Opus Claude~\citeMain{anthropic2024claude} seem to be slightly better at logical tasks like mathematics, we are not aware of any benchmark where ChatGPT was significantly outscored in tasks that involve common knowledge like HellaSwag~\citeMethod{zellers-etal-2019-hellaswag}. 

\subsection*{Variables}

To assess the actual and impersonated debate content, we measure the following variables:
\begin{itemize}
    \item \textit{Authenticity}: The likelihood that the response is an actual contribution by the speaker in a debate. This variable measures the core aspect of our study, i.e., if people believe that a statement is genuine.
    \item \textit{Coherence}: The logical flow of the response. This variable measures the internal reasoning structure of responses. 
    \item \textit{Relevance}: The extend to which the the response addresses the question. This variable measures if the responses stay on topic and convey relevant information.
    \item \textit{Content}: Whether the overall meaning of both responses is identical. This variable allows us to understand if LLM-generated responses differ from the actual responses. 
    \item \textit{Confidence}: The confidence in judging whether the response was by this speaker. This variable is used as control variable to understand if the certainty in judging debate content is affected by whether it is real or impersonated. 
    \item \textit{Familiarity}: The knowledge about a speaker from previous public appearances. This variable is used as a control variable to understand if familiarity with a speaker has an impact on the authenticity judgments. \\
\end{itemize}

\subsection*{Survey design}

We measure these variables using a survey. The survey starts with the collection of demographic data about the participants, i.e., their age, gender, country of residence within the United Kingdom, and political preference. At this time, the participants are only informed that the debate questions and responses are from the BBC show QT, but not that some responses were generated by a LLM, i.e., we use a deceptive design that rather makes participants believe they only judge actual debate content. Afterwards, the participants are randomly sorted into three tracks, such that we end up with two judgments for every data point. Each track provides a different perspective on the relationship between actual and impersonated responses. 

The goal of the first track is to collect data regarding the judgment of the authenticity, coherence, and relevance of debate responses when only a single response is shown. The participants are shown a question, a response, and the name of the speaker. The response is either the actual response by the speaker or a response we generated with an LLM, as described above. 

The second track augments this setting that the 
actual and impersonated responses are shown side-by-side: the participants see a question, the name of the speaker and both responses at the same time. Their task is to compare them with each other: which is more authentic, coherent, and relevant. Whether the actual and impersonated response is shown on the left side is randomized. Additionally, we use this comparative assessment to collect data on whether the content of the impersonated responses is the same as of the actual responses. 

The third track helps us understand different factors that could explain differences in authenticity. For this, the participants are shown a question, a response, the name of the speaker, and the short biography of the speaker. The biography is the same that we provide to the LLM as part of the user prompt. There are three populations for the statistical analysis in this track. Same as before, we show the actual speaker and the actual response, as well as the actual speaker and the impersonated response. Additionally, we also create a population in which we keep the actual response from the actual speaker, but switch the name and biography with a randomly selected different public person from our data set. The participants are then asked again to judge the authenticity of the responses, but also to rate their confidence in the judgment and their familiarity with the speaker. 

Once the participants have completed their track, they conduct an exit poll, in which we ask questions regarding their familiarity with AI and chat bots, their opinion on the use of AI in public debates, and the need for transparency and regulation in this setting. Only after this exit poll is completed, we reveal that parts of the debate responses were generated with the help of a LLM. The participants see a summary of their contribution, including which ratings they provided and which responses were actual or impersonated. Based on this new awareness of the potential of AI in debates, we repeat the exit poll to gather data on whether this affects the participants' opinions about AI in public debates. The participants can provide an (optional) free-text comment regarding their judgments from the exit poll. 


For all questions in the three tracks and the exit poll, we use a five-point Likert scales such that the middle point of the scale is neutral. The full questions and the scales can be found in the supplemental material. The survey is designed so that every participant judges eight different responses. For the first and third track, this means that each participant judges eight data points, as we only show a single response. We use rejection sampling to ensure each question/speaker pair only appears once, i.e., it is not possible that a judges both the actual and the impersonated response from a speaker to a question. For the second track, this means that the participant judges four pairs of actual questions and impersonated responses.

\subsection*{Qualitative analysis}

We use inductive coding~\citeMethod{Thomas2006} and have one author  assign one or more codes to the free-text answers from the survey. The codes are aimed to capture the intent of the free-text answer, e.g., convey reason for changes in the exit poll, or observations regarding the impersonated content the participants found striking. This is initially done for twenty of the answers, at which point the coding is checked by and discussed with a second author, resulting in an agreed-upon coding strategy. The first author then continues to code the remainder of the data. Upon completion of this coding, the second author again checks all codes and discusses the coding to achieve agreement in the same manner as for the initial set of codes. We then conduct one round of axial coding~\citeMethod{corbin2014basics} to group related codes into categories. Same as above, the axial coding is initially conducted by one author, then checked by and discussed with a second author to achieve agreement.

\subsection*{Survey participants}

Since the debate content we study originates from a popular British topical debate program, we recruited a representative sample of British citizens above the age of 18. We used Prolific for this recruitment and participants received a participation fee for compensation. Participants were informed about the purpose of the study and consent for participation was obtained. We recruited a total of 948 participants who were split up randomly into the three tracks so that we have at least two judgments for each of the 520 question/response pairs (actual responses, impersonated responses, and actual responses with random speakers).

\subsection*{Linguistic structure}

We also analyse the impersonation from a linguistic perspective by comparing discourse-related linguistic markers measured on the actual and impersonated responses. This allows us (1) to understand if the responses share properties in the linguistic surface and (2) whether the language is related to the human judgements in terms of authenticity, relevance and coherence. To this end, we measure the following linguistic structures. 
\begin{itemize}
    \item \textit{Syntactic complexity}: Syntactic complexity in terms of the mean number of conjuncts, clausal modifiers of nouns, adverbial clause modifier, clausal complements, clausal subjects and parataxis per sentence is a useful tool to understand the complexity of the language.\citeMethod{weiss-etal-2019-computationally}
    \item \textit{Modals}: The number of modal constructions (e.g., `definitely', `potentially') per sentence signals the stance of the speaker towards the utterance. 
    \item \textit{Nominalizations}: The number of nominalizations per sentence is associated with the 
    complexity of the language.\citeMethod{siskou2022measuring}
    \item \textit{Discourse markers}: The number of discourse markers (e.g. first, moreover) per sentence is associated with the coherence of texts and the use of clear argumentation structure.\citeMethod{LENK1998245}
    \item \textit{Epistemic markers}: The number of epistemic markers (e.g., `I think', `in my opinion') 
    indicates a commitment of the speaker to the message they convey.
    \item \textit{Lexical diversity}: The lexical diversity measured with MTLD gives us a perspective on how the diverse the used vocabulary is. \citeMethod{mccarthy2010mtld}
    \item \textit{Lexical overlap}: For lexical overlap between the question and the response we measure the percentage of words from the question (excluding stop words) that also appear in the response. This provides us with an approximation regarding the influence of the question on the response. 
\end{itemize}

\subsection*{Statistical analysis}

We measure the inter-rater reliability with Cronbach's $\alpha$~\citeMethod{Cronbach1951} between the two judgments for authenticity, coherence, relevance, and content. Additionally, we report the pair-wise differences between the two participants to understand which disagreements our participants have. We exclude confidence and familiarity because we cannot expect agreement regarding a subjective self-reflection. For the subsequent statistical analysis, we map the Likert scales to the integers [-2, -1, 0, 1, 2] and compute the average rating between the two judgments for the same data point. Since the variables from our survey are based on Likert scales, we use non-parametric rank-based statistical tests. 

For the first track, we assess the difference in the variables authenticity, coherence, and relevance between the actual responses and the impersonated responses. The track has a between-subjects design (i.e., different participants judge the actual and impersonated responses) with data that is paired by the question and the speaker. Consequently, we use a two-sided Wilcoxon signed rank test~\citeMethod{Wilcoxon1945} to determine if the difference between both populations is significant.

For the second track, we post-process the data such that the actual response is always on the left and the impersonated response is always on the right. The track uses a within-subjects design (i.e., a participant judges both the actual and impersonated response). We conduct a two-sided one-sample Wilcoxon signed rank test to determine if the judgments regarding authenticity, coherence, relevance, and content are significantly different from zero. For authenticity, coherence, and relevance, a significant tendency towards negative values means that the participants favour the actual responses, a significant tendency towards positive values means that the generated, impersonated responses are favoured. For the content, a significant positive value means that the contents are similar, a negative value means that the impersonated content is different from the actual content by the speaker. 

With the data for the third track, we assess if the authenticity and confidence depend on whether the speaker is real, random, or impersonated, i.e., we have three populations. The track has a between-subjects design where the populations are paired by the question and the actual  speaker. We use a Friedman test~\citeMethod{friedman1937} to test if there is any difference between the three populations with a Bonferroni-Dunn post-hoc test based on pair-wise two-sided Wilcoxon signed rank tests to determine which differences between pairs are significant. Additionally, we use the familiarity judgments to understand how this affects the authenticity. For this, we conduct a subgroup analysis where we split the ratings into those where the familiarity is less than 0 (i.e., ratings where the participants are not at least fairly familiar with the speaker) and judgments with a familiarity greater than or equal to 0. For the latter subgroup we do not have paired data anymore.
The reason for this is that we have independent raters for the three populations. For instance, the raters for the responses attribute to the actual speakers may be familiar with different speakers than the raters for the impersonated responses, leading to different subgroups. Consequently, we use a Kruskal-Wallis test~\citeMethod{Kruskal1952} with Bonferroni-Dunn post-hoc tests based on pair-wise two-sided  Wilcoxon–Mann–Whitney tests~\citeMethod{Mann1947}. 

The statistical analysis of the linguistic surface markers is similar to the analysis of the first track since we also have two populations for each linguistic marker that describe the actual and impersonated responses. Since the data from the linguistic markers does not follow a normal distribution (visual analysis of the distribution in Figure~\ref{fig:lingustic_surface} shows, e.g., long tails), we also use non-parametric tests, namely two-sided Wilcoxon signed rank tests to determine if differences for each of the seven linguistic markers are significant. 

Thus, we conduct three statistical tests with the data from the first track, four with the data from the second track, three with the data from third track, and seven for the linguistic markers, i.e., a total of 17 tests. We use a conservative approach based on Bonferroni correction~\citeMethod{bonferroni1936teoria} to account for multiple tests and consider results as significant if the p-value of a test is less than $\alpha=\frac{0.05}{17} \approx 0.003$. Based on the large size of our our populations with 520 question/response pairs and assuming that we observe differences of 0.5 points (i.e., half a step on the Likert scales), we compute the expected statistical power as $\beta=1$. Consequently, in case there are differences of 0.5 or larger in judgment, this should always be picked up by our tests and if there are no differences in judgments, we only have a 5\% chance that we find a difference that is not there. While our data is not perfectly normal, it also does not have severe outliers or multimodalities, so we prefer the clear interpretation of the arithmetic mean (M) and standard deviation (SD) to report statistical markers for populations, as well as Cohen’s $d$~\citeMethod{cohen2013statistical} for the effect size, over the slightly more appropriate, but less accessible non-parametric statistics and effect size measures. We use violin plots to visualize the distribution of the data based on a kernel density estimation of the underlying probability distribution. The violins include miniature box plots that depict the median, upper and lower quartile and the whiskers defined as the largest/smallest observed value at most 1.5 times the inter-quartile range away from the upper/lower quartile. Additionally, we use stacked bar charts to depict ratios of Likert scale items, where appropriate. 

Our statistical analysis of the data is mostly implemented in Python. We use pandas 2.2.2 and numpy 1.26.4 for the processing of data, pingouin 0.5.4 for the calculation of Cronbach’s $\alpha$, scipy 1.13.0 for the statistical tests, and seaborn 0.13.2 for the generation of plots. We compute the statistical power with the R package mkpower 0.9. 

\bibliographystyleMethod{unsrt}
\bibliographyMethod{custom.bib}

\section*{Author information}

\subsection*{Contributions}

S.H. and A.HJ. provided the initial idea for the study. All authors contributed to the design of the survey. A.T implemented and ran the survey, as well as engineering a suitable prompt to generate debate responses. Z.K. implemented analysis of the linguistic surface and conducted the initial round of inductive coding for the free-text answers and checked the outcome of the axial coding of categories. S.H. designed and implemented the statistical analysis, checked the results of the inductive coding and performed the axial coding of categories. S.H. wrote the main text, the methods, and the supplementary information. All authors gave critical feedback on the main text and the methods. 

\subsection*{Corresponding author}

Correspondence to Steffen Herbold.

\section*{Ethics declarations}

\subsection*{Competing interests}

The authors declare no competing interests.

\section*{Data availability}

The datasets generated during and/or analysed during the current study are available online at \url{https://github.com/aieng-lab/replication-kit-qtgpt-study} and \url{https://doi.org/10.5281/zenodo.12698364}.

\section*{Code availability}

All materials are available online in form of a replication package that contains the data and the analysis code at \url{https://github.com/aieng-lab/replication-kit-qtgpt-study} and \url{https://doi.org/10.5281/zenodo.12698364}.

\section*{Extended data}

\FloatBarrier

\begin{figure}[h]
    \centering
    \includegraphics[width=0.3\textwidth]{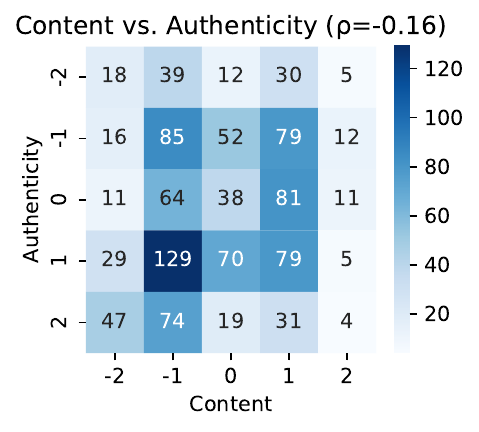}
    \caption{Judgements of the relationsship between the authenticity and content when the actual response and the ChatGPT-generated response were shown side by side. The heatmap visualizes the counts of the different rating combinations. The reported statistical marker is the correlation measured with Spearman's $\rho$.}
    \label{fig:extended-data-content}
\end{figure}

\begin{figure}[h]
    \centering
    \includegraphics[width=\textwidth]{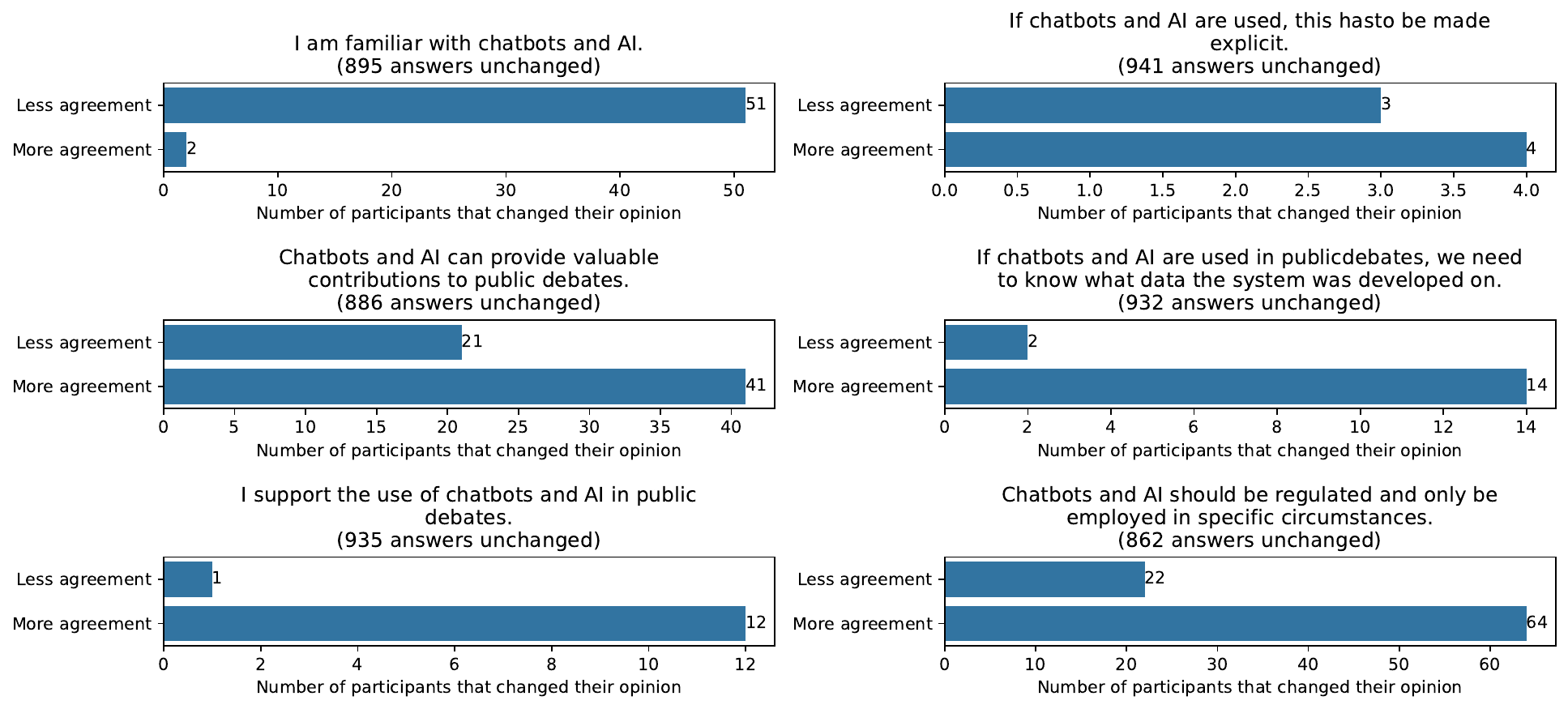}
    \caption{Changes in the participants opinions after it was revealed which responses were AI generated. The bar charts depict the counts of changes per question.}
    \label{fig:exit-poll-changes}
\end{figure}

\begin{figure}[h]
    \centering
    \includegraphics[width=0.75\textwidth]{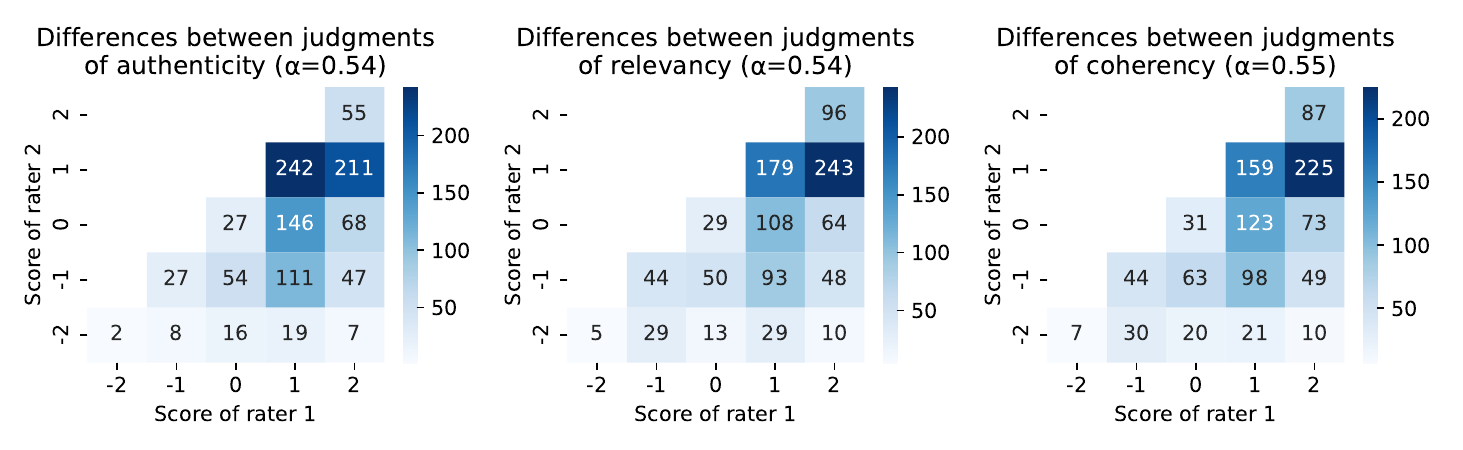}
    \caption{Inter-rater reliability when only a debate question, the name of the speaker, and either the ChatGPT-generated response or the response of the actual speaker were shown. The heatmap visualizes the counts of the different rating combinations. The reported statistical marker is Cronbach's $\alpha$.}
    \label{fig:track-1-reliability}
\end{figure}

\begin{figure}[h]
    \centering
    \includegraphics[width=0.25\textwidth]{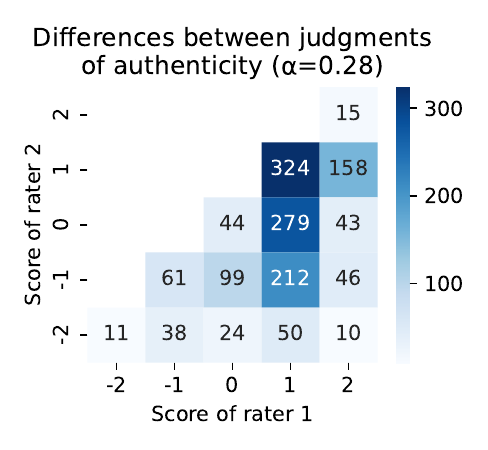}
    \caption{Inter-rater reliability when a debate question with either the response and biography from the actual speaker, the ChatGPT-generated response and the biography of the actual speaker, or the response from the actual speaker but the biography of a random public person were shown. The heatmap visualizes the counts of the different rating combinations. The reported statistical marker is Cronbach's $\alpha$.}
    \label{fig:track-2-reliability}
\end{figure}

\begin{figure}[h]
\centering
\includegraphics[width=\textwidth]{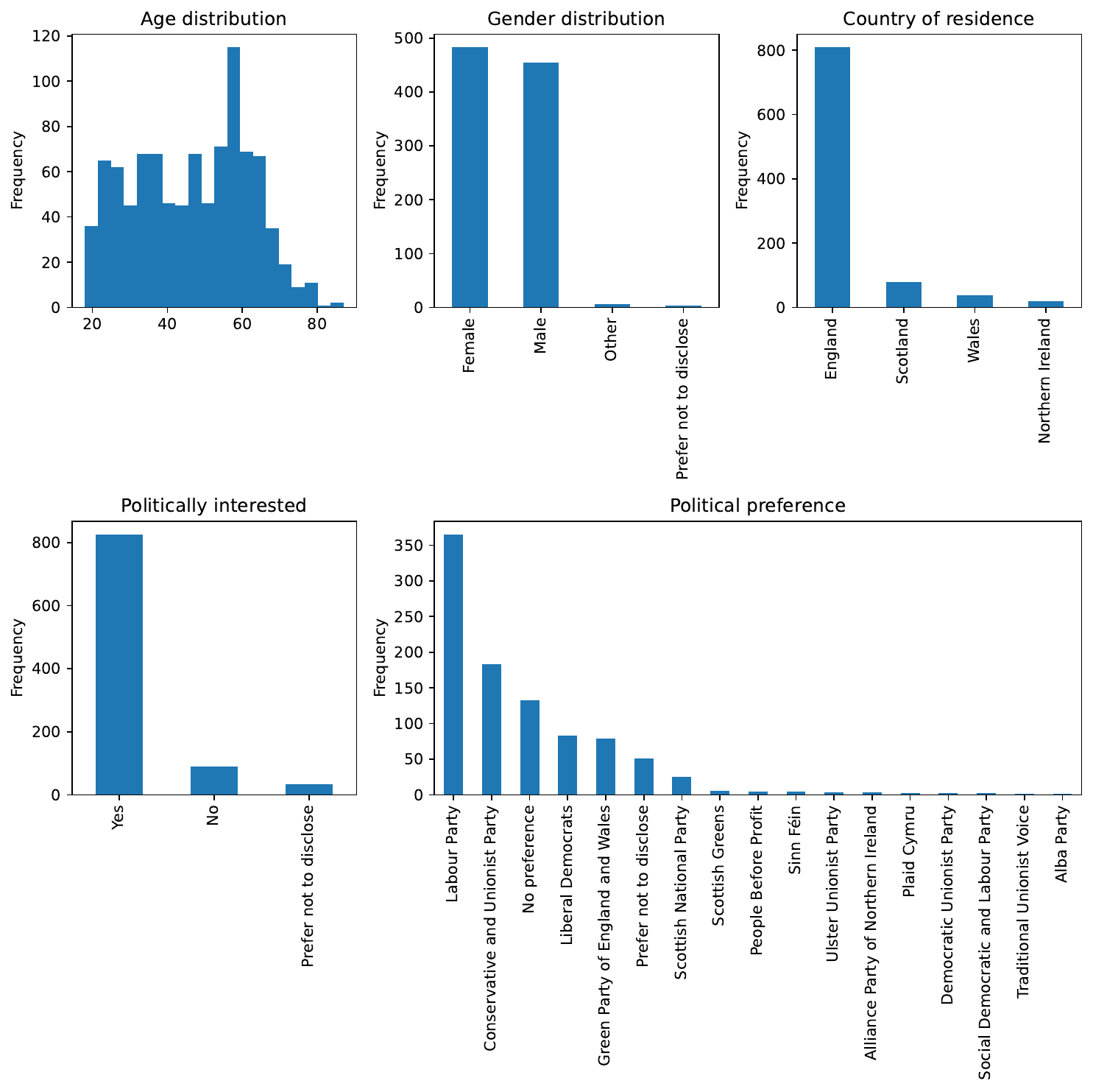}
\caption{Demographics data of our survey participants. The histogram for the age shows the distribution of the different age categories. The bar charts for the other aspects show the counts per category.}
\label{fig:demographics}
\end{figure}

\begin{figure}[h]
\centering
\includegraphics[width=0.4\textwidth]{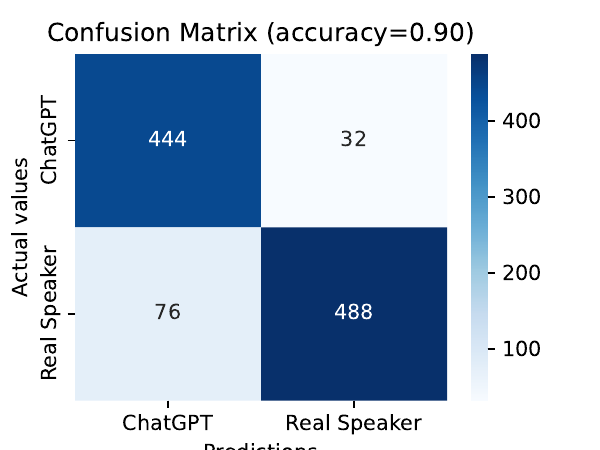}
\caption{Confusion matrix of the automated detection of the impersonated debate responses. The accuracy is the percentage of correct results, the cells depict the counts of the respective combinations.}
\label{fig:confusion-matrix}
\end{figure}

\FloatBarrier

\section*{Supplemental material}

\subsection*{Additional details for the survey}

This supplemental material provides additional details about the survey of British citizens to judge the actual and ChatGPT-generated debate content. 

\subsubsection*{Survey questions and scales}

\begin{table}[h]
\centering
\begin{tabular}{lp{8cm}}
\textbf{Question} & \textbf{Scale} \\
\hline
Age & Age in years \\
Gender & Male, Female, Other, Prefer not to disclose \\
Country of Residence & England, Scotland, Wales, Northern Ireland \\
Politically interested & Yes, No, Prefer not to disclose \\
Political preference & Conservative Unionist Party, Labour Party, Scottish National Party, Liberal Democrats, Democratic Unionist Party, Sinn Fein, Plaid Cymru, Social Democratic and Labour Party, Alba Party, Green Party of England and Wales, Alliance Party of Northern Ireland, Ulster Unionist Party, Scottish Greens, Traditional Unionist Voice, People Before Profit, No Preference, Prefer not to disclose \\
\end{tabular}
\caption{Demographic data collected as part of the survey.}
\label{tbl:questions_demographics}
\end{table}

\begin{table}[h]
\centering
\begin{tabular}{p{4.2cm}p{7cm}}
\textbf{Question} & \textbf{Scale} \\
\hline
Q1.1: The response to the question is authentic. &  Strongly disagree, Disagree, Neutral, Agree, Strongly Agree \\
Q1.2: The response to the question is coherent. & Strongly disagree, Disagree, Neutral, Agree, Strongly Agree \\
Q1.3: The response to the question is relevant. & Strongly disagree, Disagree, Neutral, Agree, Strongly Agree \\
\end{tabular}
\caption{Questions and Likert scales used for the first track, where a single question, the speaker name, and either the ChatGPT-generated or the actual response were shown.}
\label{tbl:questions_track1}
\end{table}

\begin{table}[h]
\centering
\begin{tabular}{p{4.2cm}p{7cm}}
\textbf{Question} & \textbf{Scale} \\
\hline
Q2.1: Which of the responses is more authentic? &  Left is significantly more authentic, Left more authentic, Both equally authentic, Right more authentic, Right significantly more authentic \\
Q2.2: Which of the responses is more relevant to the question? & Left is significantly more relevant, Left is more relevant, Both equally relevant, Right more relevant, Right significantly more relevant \\
Q2.3: Which of the responses is more coherent? & Left is significantly more coherent, Left is more coherent, Both equally coherent, Right more coherent, Right significantly more coherent \\
Q2.4: Both answers are similar in content. & Strongly disagree, Disagree, Neutral, Agree, Strongly Agree \\
\end{tabular}
\caption{Questions and Likert scales used for the second track, where either the response and biography from the actual speaker, the ChatGPT-generated response and the biography of the actual speaker, or the response from the actual speaker but the name and biography of a random public person were shown.}
\label{tbl:questions_track2}
\end{table}

\begin{table}[h]
\centering
\begin{tabular}{p{4.2cm}p{7cm}}
\textbf{Question} & \textbf{Scale} \\
\hline
Q3.1: The response to the question came from the speaker described above. &  Strongly disagree, Disagree, Neutral, Agree, Strongly Agree \\
Q3.2: I am confident in my previous judgment. & Strongly disagree, Disagree, Neutral, Agree, Strongly Agree \\
Q3.3: I am familiar with the speaker described above. & I am not familiar with the speaker, My familiarity with the speaker is limited, I am fairly familiar with the speaker, I am somewhat familiar with the speaker, I am familiar with the speaker \\
\end{tabular}
\caption{Questions and Likert scales used for the third track, where a single question, the speaker name were shown together with the ChatGPT-generated and actual response side-by-side.}
\label{tbl:questions_track3}
\end{table}

\begin{table}[h]
\centering
\begin{tabular}{p{4.2cm}p{7cm}}
\textbf{Question} & \textbf{Scale} \\
\hline
E1: I am familiar with chatbots and AI. &  Strongly disagree, Disagree, Neutral, Agree, Strongly Agree \\
E2: Chatbots and AI can provide valuable contributions to public debates. & Strongly disagree, Disagree, Neutral, Agree, Strongly Agree \\
E3: I support the use of chatbots and AI in public debates. & Strongly disagree, Disagree, Neutral, Agree, Strongly Agree \\
E4: If chatbots and AI are used, this has to be made explicit. & Strongly disagree, Disagree, Neutral, Agree, Strongly Agree \\
E5: If chatbots and AI are used in public debates, we need to know what data the system was developed on. & Strongly disagree, Disagree, Neutral, Agree, Strongly Agree \\
E6: Chatbots and AI should be regulated and only be employed in specific circumstances. & Strongly disagree, Disagree, Neutral, Agree, Strongly Agree \\
\end{tabular}
\caption{Questions and Likert scales used for the exit poll.}
\label{tbl:questions_exitpoll}
\end{table}

\subsubsection*{Codes and categories of free-text answers}

\begin{itemize}
\item \textit{AI as support for humans in debates}: ai\_as\_tool, ai\_use\_training\_tool
\item \textit{AI has bad quality answers}: ai\_bad\_quality, ai\_bad\_quality\_authentic,\\ ai\_bad\_quality\_confusing\_statements, ai\_bad\_quality\_sentences
\item \textit{AI better coherence due to debate setting}: ai\_better\_coherence\_expected, ai\_better\_quality\_coherence\_expected
\item \textit{AI answers better than humans}: ai\_better\_quality, ai\_better\_quality\_accuracy, ai\_better\_quality\_argumentation, ai\_better\_quality\_articulate, ai\_better\_quality\_authentic, ai\_better\_quality\_authenticity, ai\_better\_quality\_clearer, ai\_better\_quality\_coherence, ai\_better\_quality\_coherent, ai\_better\_quality\_convincingness, ai\_better\_quality\_detailed, ai\_better\_quality\_evidence\_based, ai\_better\_quality\_flow, ai\_better\_quality\_fluency, ai\_better\_quality\_grammar, ai\_better\_quality\_honesty, ai\_better\_quality\_informative, ai\_better\_quality\_less\_emotion, ai\_better\_quality\_reasoning, ai\_better\_quality\_relavance, ai\_better\_quality\_relevance, ai\_better\_quality\_structure, ai\_better\_quality\_understanding, ai\_better\_quality\_usefulness
\item \textit{AI use for debates should be responsible and regulated}: ai\_data\_source, ai\_use\_disclosed, ai\_use\_regulated, ai\_use\_regulated\_limited, ai\_use\_responsibly, ethical\_concerns
\item \textit{AI can be dangerous and misused}: ai\_fact\_correctness, ai\_misuse, ai\_replace\_humans, ai\_use\_caution, ai\_use\_danger, ai\_use\_deceive, ai\_use\_misuse
\item \textit{AI has good quality answers}: ai\_good\_balanced, ai\_good\_quality, ai\_good\_quality\_accuracy, ai\_good\_quality\_adapts\_to\_new\_topics, ai\_good\_quality\_authenticity, ai\_good\_quality\_balanced, ai\_good\_quality\_clarity, ai\_good\_quality\_coherence, ai\_good\_quality\_considered, ai\_good\_quality\_convincing, ai\_good\_quality\_convincingness, ai\_good\_quality\_detailed, ai\_good\_quality\_fluency, ai\_good\_quality\_reasoning, ai\_good\_quality\_relevance, ai\_good\_quality\_sentence\_structure, ai\_quality\_good\_summarization
\item \textit{AI only imitates humans}: ai\_use\_imitates
\item \textit{AI is indistinguishable from humans}: ai\_indistinguishable
\item \textit{Less familiar with AI than expected}: ai\_not\_as\_familiar
\item \textit{AI use suspected}: ai\_too\_coherent, suspected\_ai\_involvement
\item \textit{AI should not be used for debates}: ai\_use\_defeats\_purpose, ai\_use\_limited, ai\_use\_no
\item \textit{Undecided about use of AI in debates}: ai\_use\_maybe, ai\_use\_unclear, ai\_use\_undecided, ai\_use\_undecided'
\item \textit{Support use in debates}: ai\_use\_yes, ai\_valuable\_contribution, ai\_valuable\_contribution, ai\_valuable\_contributions
\item \textit{AI answers worse than humans}: ai\_worse\_quality\_authenticity, ai\_worse\_quality\_novelty
\item \textit{More information could help AI detection}: awareness\_could\_help, context\_to\_distinguish
\item \textit{Other}: bad\_quality\_does\_not\_matter, change, familiar\_with\_QT, made\_mistake\_in\_survey, nature\_of\_ai, no\_change, satisfied\_with\_responses, study\_encourages\_caution, survey\_structure\_comment, time\_tracker\_issue, unfamiliar\_with\_speakers
\item \textit{Negative emotions}: emotion\_alarmed, emotion\_confusion, emotion\_deceived, emotion\_dismay, emotion\_fear, emotion\_shock, emotion\_unhappiness, emotion\_worry
\item \textit{Positive emotions}: emotion\_amazement, emotion\_fascinated, emotion\_impressed, emotion\_surprise, emotion\_surprised
\item \textit{Humans have bad quality answers}: human\_quality\_bad, human\_quality\_bad\_coherence, people\_lie, quality\_human\_bad
\item \textit{ Human can express opinions}: human\_quality\_good\_own\_opinion
\item \textit{Human answers better than AI answers}: preferred\_real
\end{itemize}


\end{document}